\documentclass[11pt]{article}
\usepackage{acl2015}
\usepackage{times}
\usepackage{url}
\usepackage{latexsym}
\usepackage[utf8]{inputenc}
\usepackage[none]{hyphenat}
\usepackage{multirow}
\usepackage{amsmath}
\usepackage{epsfig}
\usepackage[nolist,nohyperlinks]{acronym}
\title{LIA-RAG: a system based on graphs and divergence of probabilities applied to Speech-To-Text Summarization}

\author{Elvys Linhares Pontes \\
  Universidade Federal do \\
  Ceará, Campus Sobral  \\
  Fortaleza, Brasil \\
  {\tt elvyslpontes} \\
  {\tt @gmail.com} \\\And
  Juan-Manuel Torres-Moreno \\
  LIA / UAPV \\
  BP 1228, 84911 Cedex 9 \\
  Avignon,  France \\
  and \'Ecole Polytechnique \\
  de Montréal (Canada)\\
  {\tt juan-manuel.torres} \\
  {\tt @univ-avignon.fr}\\\And
  Andréa Carneiro Linhares \\
  Universidade Federal do \\
  Ceará, Campus Sobral \\
  Fortaleza, Brasil \\
  and LIA / UAPV France \\
  {\tt andrea.linhares@ufc.br} \\}

\date{}

\begin{document}

\begin{acronym}{}
\acro{ATS}{Automatic Text Summarization}
\acro{CAS}{Chemical Abstracts Service}
\acro{CST}{Cross-document Structure Theory}
\acro{JS}{Jensen-Shannon}
\acro{KL}{Kullback-Leibler}
\acro{MRS}{Multi-document Rhetorical Structure}
\acro{NLP}{Natural Language Processing}
\acro{ROUGE}{Recall-Oriented Understudy for Gisting Evaluation}
\acro{RAG}{\textsl{Résumeur Audio-texte à base de Graphes}}
\end{acronym}

\maketitle
\begin{abstract}
This paper aims to introduces a new algorithm for automatic speech-to-text summarization based on statistical divergences of probabilities and graphs. The input is a text from speech conversations with noise, and the output a compact text summary. Our results, on the pilot task CCCS Multiling 2015 French corpus are very encouraging.
\end{abstract}


\section{Introduction}

Nowadays, a lot of information is daily generated. It is necessary to have available memory storage because each datum must be processed and the information contained therein analyzed. The manual analysis is impossible because it is necessary a huge number of persons to analyze this information in an available time. The summary is a short text with main ideas of original text \cite{torres2014automatic} and reduces the read time to analyze these data.

Audio is widely used in daily life on the radio and on the internet, in news, interviews and conversations. A Call Centre Conversation creates a lot of conversations every day. These centers has issues and tasks. It is essential the control of the discussed topics and the results obtained by customers in these calls. One way to analyze and accelerate the data processing is speech summarization, that is different from traditional text summarization because there are other problems in these texts as speech errors, sentences of different sizes and colloquialisms.

\textsl{``Multiling is a community-driven initiative for benchmarking multilingual summarization systems, nurturing further research, and pushing the state-of-the-art in the area''}\footnote{\url{http://multiling.iit.demokritos.gr/pages/view/1517/multiling-2015-call-for-participation}}. The MultiLing 2015 initiative features the following tasks: Multilingual Multi-document Summarization, Multilingual Single-document Summarization, Online Forum Summarization and Call Centre Conversation Summarization (CCCS). The CCCS pilot task consists in \textsl{``creating systems that can analyze call centres conversations and generate written summaries reflecting why the customer is calling, how the agent answers that query, what are the steps to solve the problem and what is the resolution status of the problem"} \cite{benoit}.

We developed the LIA-RAG summarization system based on the RAG system \cite{rag:15}, coupled with some post-processing rules
in order to generate a final summary. LIA-RAG uses a graph model to analyze and verify a set of documents (e.g., the conversation transcription) for MultiLing'15 CCCS pilot task. 
LIA-RAG creates a summary computing the relevance of the words and the similarity among the sentences. The system uses a simple post-processing to improve the quality of the final summary. 

The rest of the paper is organized as follows: 
section \ref{sc:rt} describes related work on automatic summarization of texts and conversations. Sections \ref{sc:mod} and \ref{sc:rag} analyze the graph model and the system used in this work. Section \ref{sc:results} describes the results obtained for Multiling/DECODA French corpus and section \ref{sc:conc} concludes this work.

\section{Related Works}
\label{sc:rt}

\ac{ATS} aims to creates a summary containing the main ideas of a textual document \cite{mani:mayburi:99,mani:01,torres2014automatic}. The summary can be an extraction or abstraction of a single document or multi-document. The extraction process identifies the most informative sentences of a document and creates a summary by assembling of these sentences \cite{Luhn,torres2014automatic}. Extraction may be guided (by a query). In this case, the algorithm selects the most relevant information follow a particular topic. The abstraction algorithms create new (or reformulate) sentences from original texts \cite{seno1,seno2} and the extraction methods use the key sentences of texts \cite{Barzilay,torres2014automatic}.

Works about abstraction usually uses syntactic and semantic knowledge of a language to create the summary. This procedure verifies the best construction of a sentence \cite{Barzilay2}. This type of summarization uses fusion to help the review of information. \cite{seno1} proposed a method to fusion similar sentences in Brazilian Portuguese based on a symbolic and domain-independent approach. This method allows the fusion by union and by intersection of a document cluster. Fusion by union preserves the overall message of the cluster while fusion by intersection analyses the redundant information considered most important in the cluster. \cite{seno2} described how to identify common information between sentences in Brazilian Portuguese using lexical 
knowledge, syntactic and semantic rules of paraphrasing.

\cite{cstsumm} developed a summarizer system based on the CST model (Cross-document Structure Theory). The system proposed analyses redundancy and contradiction among different information sources in Brazilian Portuguese.

\cite{Barzilay2} developed a method to generate automatic summaries by identifying and synthesizing similar elements in a cluster of documents. This method creates the summary based on similarity between the sentences and topic. \cite{Barzilay} described an approach to fusion sentences through the text-to-text technique, to synthesize repeated information from multiple documents. This method uses a syntactic alignment in sentences to identify common information. After the identification step, sentences are processed and a new text is generated with the same content.

A way to calculate the similarity between sentences is to use co-occurrence of words. \cite{He} proposed a fusion method using similarity metrics, co-occurrence skip-bigram and information density to evaluate sentences and to select the most relevant ones. \cite{Hennig} developed a multi-document model to summarize by analyzing the co-occurrence of sentence-term and sentence-bigram using the \ac{JS} divergence.

Another method to obtain relevant sentences uses compression, as reported in \cite{Pitler}. Pitler uses approaches based on syntactic trees, sentences and 
discourse. \cite{Filippova} describes a multi-sentence compression method using a word-based graph.
 
The summarization by extraction does not have the same quality as the summaries produced by abstraction because it uses surface methods based on statistical calculations to verify the sentence relevance. However, the extraction is general and do not require deep analysis of the language \cite{Barzilay,sasi}.

\cite{sasi} use Graph theory concomitant with \ac{JS} divergence to create multi-document summaries by extraction. Their system describes a text model as a graph where the sentences are represented by vertices and the edges connect two similar sentences. Their approach calculates the stable set of the graph aiming creating the summary containing sentences with general information of the cluster and without redundancy. \cite{glouton} model the text as graph model and use a heuristic (greedy algorithm) to obtain the relevant sentences in the text.

The speech summarization task is more complex and it involves other problems. It is more difficult to identify utterance boundaries because it may be fragmented,  contain disfluencies and also because speech recognition introduces errors. Meetings involve multi-party conversation with overlapping speakers. The language used is informal and utterances tend to be partial, fragmentary, ungrammatical and include many ellipses and pronouns. However, the speech signal may provides additional information that emphasizes a piece of text as prosody \cite{Murray}.

\cite{Mckeown} described some ways to use a text summarization as a speech summarization. They described some work about summarization of broadcast news and meetings. \cite{Murray} analyzed extractive summarization of multiparty meetings. They described Maximal Marginal Relevance and Latent Semantic Analysis to create the summary based on prosodic and lexical features.

\section{Modeling the problem}
\label{sc:mod}

This paper aims to design a system to summarize several documents by extraction its most important sentences. Statistical techniques were used to build a language independent system. The proposed methods are based on a specific preprocessing of words, a weighting function of sentences and a bag-of-words model to represent the text content.

This model uses $K$ matrices represented by $S^{K}_{[m \times n]}$ and constructed from $K$ documents, where $m_a$ is the number of sentences and $n_a$ is the number of distinct words in the document $a$ ($a \in K$). The cell $s^{a}_{ij}$ of the matrix represents the frequency of word $j$ in the sentence $i$ ($FP_{ij}$) of the document $a$.
This stage was constructed using the libraries and algorithms from Cortex summarization system \cite{torres2002condenses,Torres-Moreno2001}.

\begin{equation}
\begin{split}
 S^{a}=\left(
\begin{array}{cccc}
s^{a}_{11} & s^{a}_{12} & \ldots & s^{a}_{1n}\\
s^{a}_{21} & s^{a}_{22} & \ldots & s^{a}_{2n}\\
\vdots & \vdots &  & \vdots \\
s^{a}_{m1} & s^{a}_{m2} & \ldots & s^{a}_{mn}\\
\end{array}
\right), a \in K\\
s^{a}_{ij}=\left\{
\begin{array}{cc}
FP_{ij}, & \textrm{if}\ \exists\ \textrm{word\ j\ in\ sentence\ i}\\
0, & \textrm{otherwise}\\
\end{array}
\right.
\end{split}
\end{equation}

\subsection{Jensen-Shannon divergence}
\label{ss_djs}

We use Jensen-Shannon (JS) divergence to measure the similarity between sentences. Let $w$ be a words’ set in P and Q. P and Q represent
the probability distribution between two objects:
two individuals sentences or a sentence and a set of sentences.
The divergence will then calculated among these two objects. The \ac{JS} divergence is symmetric and provides a stable way to measure the difference between two distributions (equation \ref{DJS}).

\begin{eqnarray}
\label{DJS}
D_{JS}(P||Q) &=& \frac{1}{2}\sum_{w \in W} \Bigg[ P_w\log{\Bigg( \frac{2 \times P_w}{P_w+Q_w} \Bigg)} \nonumber \\
  &+& Q_w\log{\Bigg( \frac{2 \times Q_w}{P_w+Q_w} \Bigg)} \Bigg]\\ \nonumber
\end{eqnarray}

The \ac{JS} divergence value ranges from $[0,\infty+)$. It is closer to zero when the distributions are similar and they differ in another case. 

In the case there is a word in a sentence that is missing in another one, a smooth (different weighting) will be used to avoid null values and have a smoother distribution \cite{smoothBook}. If a word $w$ is not present in the sentence $Q$, then the smooth is calculated by the equation \ref{smooth}, where $ \beta = 1.5 \times voc$, which $voc$ is the number of distinct words in $R$, $\gamma$ is the variable that controls the relevance of the missing word in the sentence and $N$ is the number of words in $R$ 
\cite{nenkova}.

\begin{equation}
 \label{smooth}
  Q_w = \left( \frac{P_w + \gamma}{N + \gamma \times \beta} \right)
\end{equation}

\subsection{Term Frequency-Inverse Sentence Frequency (TF-ISF)}
\label{sc_tfisf}

One way to verify the initial relevance of a word and a sentence to the text is through the TF-ISF. This metric is based on term frequency in the text and it is calculated by the equation \ref{tfisf}.

\begin{equation}
 \label{tfisf}
  tf\_isf(w) = tf(w) \times \log \left( \frac{n}{n_{w}} \right)
\end{equation}

\noindent where $tf(w)$ is frequency of term $w$, $n$ is total number of documents and $n_w$ is number of documents that contain the term $w$.

\section{The LIA-RAG system}
\label{sc:rag}

In general lines, a text consists of several sentences with different topics. The text can be divided into several groups and each of them describes one step/idea in the text. If a group is large, then it is relevant to the text. It is possible to choose the sentences of the largest group and obtains the most relevant content.

The main ideas of a text are generally analyzed and discussed several times. The vertices with higher degree have more similar sentences and then, are important to the text. However, it is not necessary to have a lot of similar sentences to be a relevant one.

\ac{RAG} is a summarizer system by sentence extraction, which selects the main sentences of a text and uses a post-processing to remove some errors and make the text more concise and compact.

\subsection{The RAG algorithm}
\label{ssc:rag_desc}

\ac{RAG} uses Graph theory and divergence metrics to calculate the similarity and to group the sentences. Initially, the system performs a filtering process to remove the brackets. Then, it performs a segmentation, filtering and stemming processes to remove stopwords and reduce the words to their roots. RAG accomplished this preprocessing and matrix transformation based on \cite{Torres-Moreno2001}. It calculates the relevance of each sentence based on TF-ISF metric (equation \ref{tfisf}) and removes the less relevant sentences.

The system creates a graph $G$ which each vertex represents a sentence previously selected. The text is analyzed and modeled as a sentence graph (vertices). Based on equation \ref{tfisf}, it calculates the similarity between sentences. If the similarity between two sentences is less than 0.16 (threshold obtained by empirical testing), then the system creates an edge between them. So, the vertices with higher degrees have the most relevant content of the text. However, some sentences may have a small degree, but they may contain important information.

RAG combines the TF-ISF and degree sentences to analyze the relevance of them. The relevance of the sentence $i$ is defined by:

\begin{equation}
\label{eq:score}
  rel(i) = degree(i) \times tf\_isf(i)
\end{equation}

\noindent where $degree(i)$ is the degree of vertex $i$ and $rel(i)$ is the relevance of the sentence $i$.
After, the system creates a summary with the higher score sentences, excluding similar (or redundant) sentences based on Dice's coefficient \cite{dice}. 

\begin{figure}[!h] 
\begin{center} 
\includegraphics[scale=0.42]{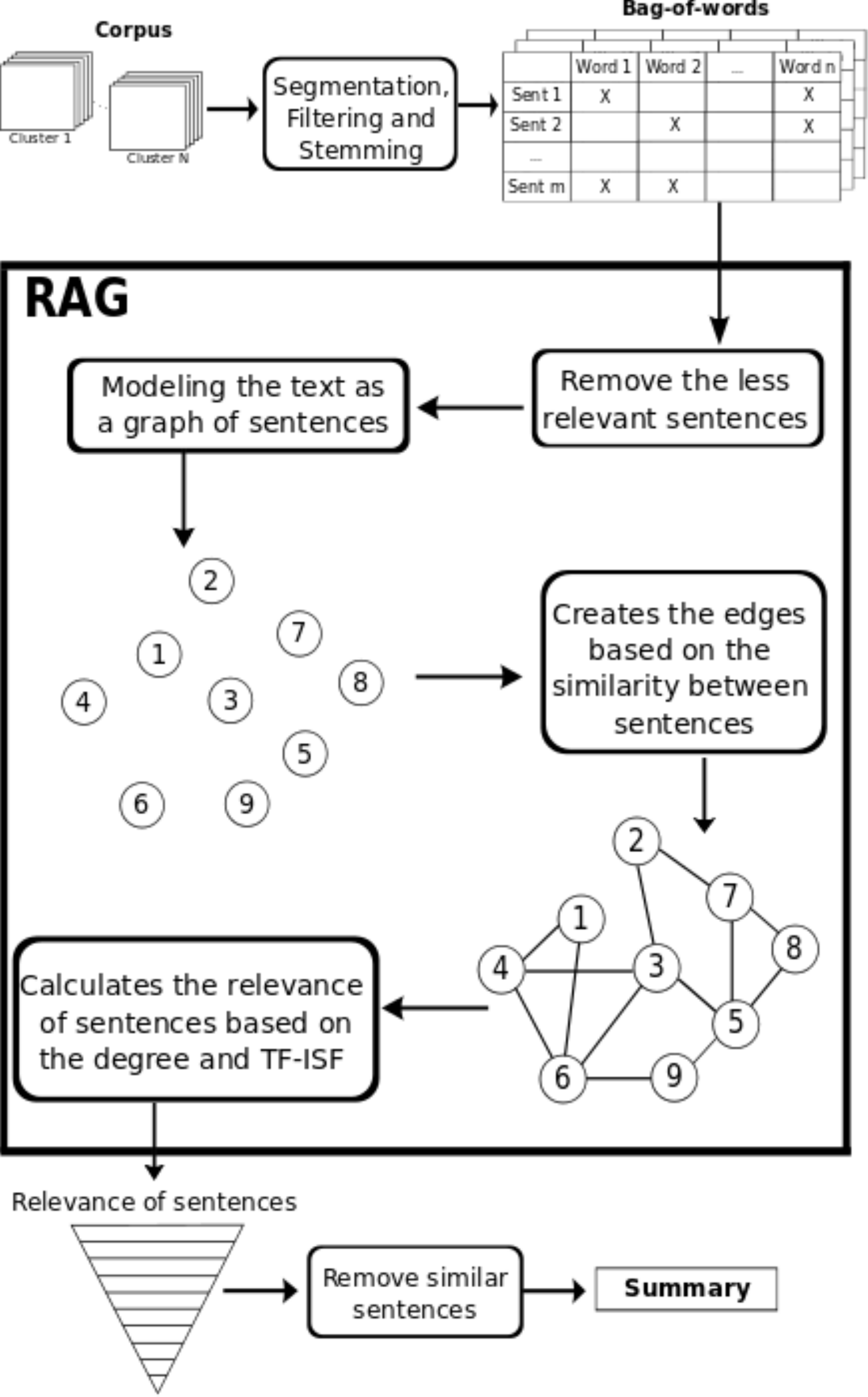}
\end{center} 
\caption{Architecture of the RAG system.} 
\label{fig:rag} 
\end{figure}

The figure \ref{fig:rag} describes the RAG system.

\subsection{LIA-RAG: RAG with a specific speech post-processing}
\label{ssc:rag_post}

The speech recognition process produces a text that contains several grammatical problems (slang, colloquialisms, expressions and speech recognition errors). An extraction summary algorithm selects the relevant sentences, however the sentences may have some grammatical problems. So, it is necessary to perform a treatment of this summary.

The main analyzed aspects in this process are: 
\begin{itemize}
\item Colloquialisms, 
\item Speech expressions and 
\item Dates. 
\end{itemize}

\begin{figure}[!h] 
\begin{center} 
\includegraphics[scale=0.38]{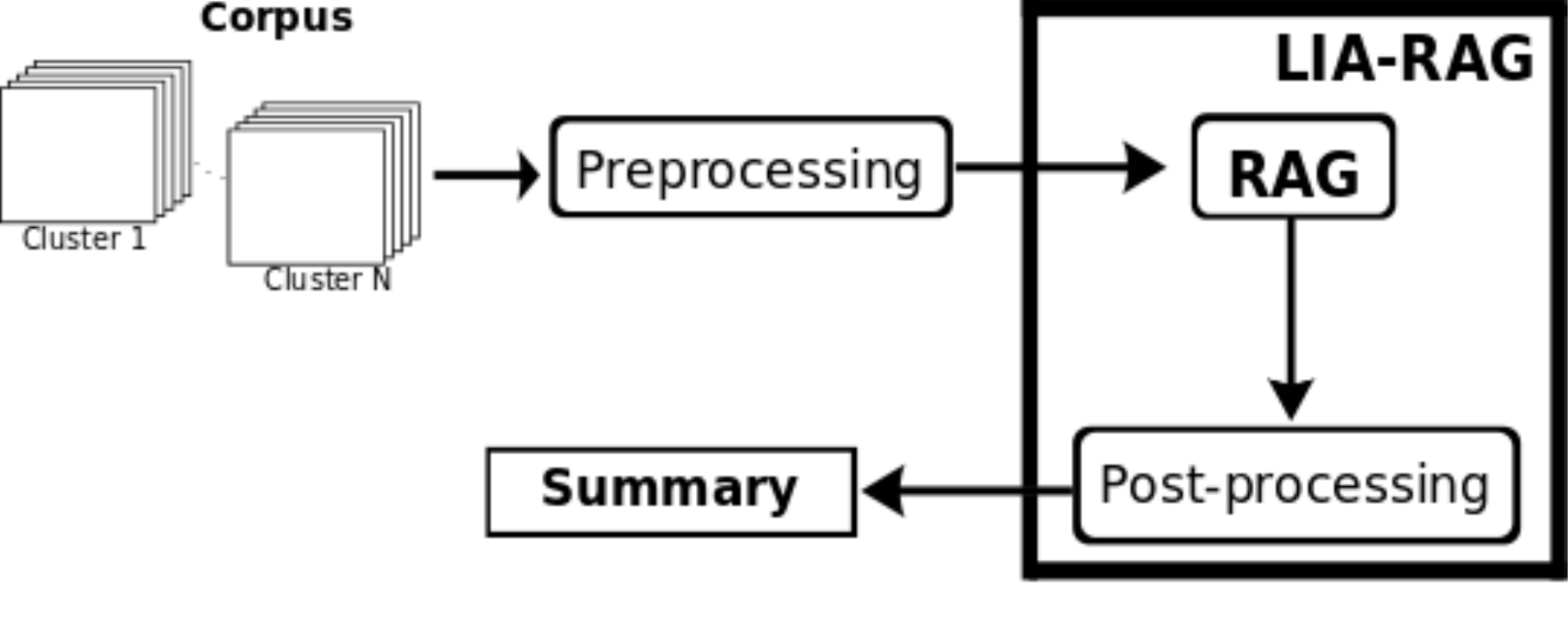}
\end{center} 
\caption{Architecture of the LIA-RAG system.} 
\label{fig:lia-rag} 
\end{figure}

LIA-RAG system receives the summary as an input.
In this input, some speech expressions are used to connect ideas or concepts in oral conversations.
LIA-RAG removes these expressions, because often they are incorrectly transcripted (a noise source). Also, the system eliminates several colloquialisms and the duplicated words. The system replaces some mistaken words by its correct form. The figure \ref{fig:lia-rag} shows the architecture of the LIA-RAG system.

\section{Results}
\label{sc:results}

The tests were carried on a computer with i5@2.6 GHz processor and 4 GB of RAM on GNU/Linux Debian 64-bit operating system. The algorithms of RAG were implemented using the Perl language.

We used the French DECODA corpus \cite{BECHET12.684}. The systems have to generate textual summaries with the main idea of each conversation belonging to the corpus. \textsl{``The conversations topics range
from itinerary and schedule requests, to lost and found, to complaints (the calls were recorded during strikes)''}\cite{benoit}. Each summary has 7\% of the number of words of each conversation transcription. We compared LIA-RAG and RAG systems with two baseline systems (random and  first lead base).

In order to evaluate the quality of the summaries, Multiling CCCS used the system \ac{ROUGE}\footnote{The options for running ROUGE 1.5.5 are -a -l 10000 -n 4 -x -2 4 -u -c 95 -r 1000 -f A -p 0.5 -t 0}, which determines the quality of an automatic summary based on the intersection of the $n$-grams of a candidate summary and the $n$-grams of a set of reference summaries. More specifically, we used ROUGE-N and ROUGE-SU measures. ROUGE-N, N $\in\ [1, 2]$. ROUGE is an $n$-gram recall measure \cite{rouge}\footnote{\url{http://www.berouge.com/Pages/default.aspx}}. The values of these metrics belongs to $[0, 1]$, 1 for the best result.

The table \ref{tb:rouge_training} shows the results obtained using the systems over the training corpus.
This corpus contains 50 conversations transcription with 23,363 words and 115 summaries. Both versions of RAG provided the best results. The RAG system identified the main sentences discussed in conversations. 
However, the errors and speech expressions decreased the informativeness. 
The post-processing of LIA-RAG allowed to improve the results. 
This process reduces errors and generates a more informative and concise summary.

\begin{table}[h]
\centering
\begin{tabular}{|c|ccc|}
\hline
 \textbf{Systems} 	& \small 
\textbf{ROUGE-1}    	& \small \textbf{ROUGE-2}    	& \small \textbf{ROUGE-4}     	\\ \hline
 \textbf{LIA-RAG:1}			&   \textbf{0.1893}  	&   \textbf{0.0628}  	& \textbf{0.0683} \\ 
 \textit{RAG}			&   \textit{0.1833}  	&   \textit{0.0614}  	& 
 \textit{0.0654} \\ 
 Base-first 	&   0.1578  	&   0.0556  	& 0.0583 \\ 
 Base-rand 		&   0.1170  	&   0.0310  	& 0.0371 \\ 
\hline
\end{tabular}
\caption{\label{tb:rouge_training} Evaluation of training corpus.}
\end{table}

The French test corpus has 100 conversations transcription with 42,130 words and 212 summaries. The ROUGE-2 official performance for the systems participating to CCCS pilot task is showed in table \ref{tb:rouge_test} \cite{benoit}. The LIA-RAG system  obtained the best results.

\begin{table}[h]
\centering
\begin{tabular}{|c|c|}
\hline
 \textbf{Systems} 	& \textbf{ROUGE-2} 	\\ \hline
 \textbf{LIA-RAG:1}		&  \textbf{0.037} \\
 NTNU:1 		&  0.035 \\ 
 NTNU:3			&  0.034 \\ 
 NTNU:2 		&  0.027 \\ 
 \hline
\end{tabular}
\caption{\label{tb:rouge_test} Evaluation of test corpus.}
\end{table}

\section{Conclusion and perspectives}
\label{sc:conc}
 
Divergence of probabilities in a graph model to extract key sentences in French speech-to-text summarization was very interesting. LIA-RAG system uses very few language resources (stopwords and stemming) and has achieved good results. 
Nevertheless, the system is easily adaptable to other languages with only some modifications in the preprocessing stage.

An interesting perspective of this work consists in the utilization of the speech TAGs markers to improve the computation of the sentences score. 
In addition, it is necessary to improve the post-processing in order to increase the quality of the final summary.
Finally, the verification of the grammaticality and readability of the extracted key sentences can help to produce more realistic abstracts.

\section*{Acknowledgments}
This project was partially founded by a scholarship from FUNCAP-CE (Brazil).

\newpage

\bibliographystyle{acl}
\bibliography{multi}

\end{document}